\documentclass[10pt,twocolumn,letterpaper]{article}

\usepackage[final]{cvpr}      
\usepackage{amsmath}
\usepackage{amssymb}
\usepackage{booktabs}
\usepackage{url}
\usepackage{mathtools} 

\usepackage{times,graphicx,amsmath,amssymb,booktabs,tabulary,multirow,overpic,xcolor}
\usepackage{colortbl}
\usepackage{makecell}
\usepackage{float}
\usepackage{bbding}

\definecolor{sh_gray}{rgb}{0.84,0.84,0.84}
\definecolor{sh_gray2}{rgb}{1,0.89,0.75}
\definecolor{color3}{rgb}{0.95,0.95,0.95}
\definecolor{color4}{rgb}{0.96,0.96,0.86}
\definecolor{color5}{rgb}{0.90,0.90,0.90}
\usepackage[pagebackref,breaklinks,colorlinks]{hyperref}
\usepackage[capitalize]{cleveref}
\usepackage{multirow}
\crefname{section}{Sec.}{Secs.}
\Crefname{section}{Section}{Sections}
\Crefname{table}{Table}{Tables}
\crefname{table}{Tab.}{Tabs.}
\usepackage{algorithm}
\usepackage{algorithmic}
\usepackage{threeparttable} 

\begin{document}

\title{Enabling Real-Time Point-of-Care Ultrasound Segmentation: A GPU-Free Deployment in Resource-Limited Settings}

\author{%
Weihao Gao$^{1,}$\thanks{Corresponding author. Email: weihaomeva@163.com} , 
$^{1}$School of Computer Science and Artificial Intelligence, Guangdong University of Education\\
}
    
\maketitle

\begin{abstract}

Ultrasound imaging is the most widely adopted medical modality globally due to its low cost and portability, yet artificial intelligence (AI) deployment remains constrained by reliance on GPU-accelerated models, creating a structural paradox where the cost of "intelligence" exceeds that of the imaging device itself. Here, we present the systematic adaptation and extensive evaluation of UltraSeg, an ultra-lightweight architecture originally developed for colonoscopic polyp segmentation, now engineered for point-of-care ultrasound (POCUS) across ten public datasets spanning six anatomical sites (breast, thyroid, kidney, carotid, fetal, and small-animal tumor). We systematically validate both variants in ultrasound domains: UltraSeg-130K (0.13M parameters) achieves 89.7 FPS on single-core CPUs and 34.8 FPS on a refurbished mobile device, while UltraSeg-500K (0.5M parameters) delivers 44.6 FPS on CPU and 16.1 FPS on mobile device. UltraSeg-500K matches or exceeds the Dice performance of the 31M-parameter UNet and approaches 105M-parameter TransUNet in average performance, with superior zero-shot cross-dataset generalization on external validation sets (UDIAT, DDTI). By enabling clinical-grade segmentation without GPU dependency, this work brings AI costs in line with ultrasound accessibility, making advanced diagnostics available in resource-limited settings.

\end{abstract}

\section{Introduction}
\label{sec:introduction}

Ultrasound imaging has emerged as the most widely adopted medical imaging modality globally, attributed to its non-invasive nature, minimal cost, and absence of ionizing radiation. It is deployed across diverse clinical settings, from tertiary hospital emergency departments to remote primary care clinics in resource-limited regions. Distinct from centralized imaging modalities (e.g., CT or MRI), ultrasound operates within a decentralized, patient-centric workflow: fetal screening, thyroid assessment, breast examination, and vascular evaluation are frequently conducted across disparate clinical units, by operators with varying expertise levels, using portable devices that transition between multiple examination rooms. This inherent mobility and distributed deployment model establish ultrasound as the quintessential Point-of-Care Ultrasound (POCUS) technology, particularly suited for bedside diagnostics in diverse healthcare tiers~\cite{debiasio2023point,sorensen2019point,venkatayogi2023seeing,frija2023paradigm}.

The clinical value of ultrasound examinations stems largely from its negligible per-scan cost and high accessibility, establishing it as the preferred imaging screening tool in developing countries and resource-limited settings. However, a persistent blind spot exists in global health policy: ultrasound penetration remains critically insufficient in primary care facilities within low- and middle-income countries (LMICs). Data from a survey across 10 countries indicate that only 1.2\% of basic primary care facilities and 3.1\% of advanced primary care facilities are equipped with ultrasound devices~\cite{yadav2021availability}. More critically, even where equipment is available, a substantial chasm persists between "having a device" and "using it effectively". A global survey of 241 frontline healthcare providers across 62 LMICs revealed that although 88\% of respondents had access to ultrasound equipment at their institutions, competition for device usage, high maintenance costs, and inadequate specialized training remain primary barriers to clinical utilization~\cite{ginsburg2023survey}.

This implementation gap extends beyond resource-limited settings. Even in high-income countries such as the United States, systematic training in ultrasound skills faces structural challenges: a national survey demonstrated that while 57\% of MD-granting medical schools have implemented POCUS curricula, faculty shortages and insufficient time prevent the vast majority from establishing longitudinal training programs~\cite{russell2022state}. This suggests that the value of AI-assisted diagnosis lies not merely in substituting scarce experts, but rather in alleviating the universal global bottleneck in ultrasound training.

The inherent operator-dependent nature of ultrasound further exacerbates implementation challenges. Image quality is profoundly influenced by operator technique, probe angulation, and anatomical knowledge, while lesion segmentation and measurement demand extensive clinical experience, rendering its learning curve considerably steeper than interpretation of CT or MRI. A nationwide cross-sectional study encompassing 5,460 ultrasound departments across Chinese hospitals revealed that even within the secondary healthcare system of a middle-income country, the ratio of sonologists to patients is merely 2.25 per 10,000, with average diagnostic accuracy for breast ultrasound at only 73.64\%, declining significantly in lower-tier hospitals or economically disadvantaged regions~\cite{gao2022breast}. This reliance on highly skilled sonographers clashes sharply with the severe shortage of experienced imaging physicians in reality. Primary care institutions often must rely on inadequately trained general practitioners or midwives for preliminary screening, resulting in compromised diagnostic consistency and accuracy. It is precisely these characteristics, decentralized deployment, inherent mobility, and heavy reliance on operator proficiency, that exacerbate the implementation barriers for digital medicine technologies in real-world clinical environments.

With the advancement of deep learning, artificial intelligence has assumed an increasingly pivotal role in clinical ultrasound diagnostics, particularly in facilitating precise lesion segmentation~\cite{alsharid2025public,kucs2024medsegbench}. Accurate lesion segmentation not only provides reliable anatomical delineation and quantitative features (e.g., size, morphology) to inform clinical decision-making with objective evidence, but also enhances diagnostic accuracy. More importantly, such AI-assisted systems can effectively reduce inter-observer variability, enabling less experienced non-expert physicians to achieve diagnostic performance comparable to specialists, thereby mitigating quality disparities arising from uneven distribution of medical resources~\cite{song2024survey,christiansen2025international}.

However, mainstream deep learning segmentation models currently face substantial deployment bottlenecks. Specialized architectures (e.g., nnUNet, TransUNet) enable fully automated segmentation yet typically comprise $>$30M parameters with computational demands reaching hundreds of GFLOPs~\cite{chen2024transunet,isensee2021nnu}. Segmentation foundation models such as SAM or MedSAM, while possessing cross-domain generalization capabilities, rely on interactive prompts (e.g., bounding boxes or point annotations) and cannot perform fully automated inference, increasing operational complexity for clinicians~\cite{zhang2025embedded,ma2024segment,kirillov2023segment}. Moreover, these models maintain parameter counts ranging from tens to hundreds of millions. Both approaches necessitate dedicated GPU support, with hardware deployment costs substantially exceeding those of the ultrasound devices themselves.

This structural imbalance creates a critical barrier: examination costs are minimal, yet AI deployment costs are prohibitive. This contradiction severely constrains the democratization of artificial intelligence in ultrasound, particularly in underserved regions, fiscally constrained primary care institutions, remote areas, or mobile healthcare scenarios lacking professional IT maintenance. When AI diagnostic tools remain restricted to large medical centers with robust digital infrastructure, the vast patient populations who would most benefit from screening with this relatively low-cost imaging modality are paradoxically excluded from technological dividends, exacerbating health inequity. When AI-assisted systems require expensive GPU workstations or stable cloud connectivity, dispersed ultrasound devices cannot readily acquire intelligent capabilities, creating a paradox of "hardware accessible yet intelligence inaccessible."

To resolve this structural mismatch between AI computational cost and ultrasound accessibility, we present UltraSeg, an ultra-lightweight framework originally developed by our group for real-time colonoscopic polyp segmentation~\cite{gao2026enabling}, now systematically adapted and clinically validated for decentralized POCUS scenarios. This study establishes the systematic clinical efficacy of this parameter-scalable architecture across heterogeneous POCUS domains on ten public datasets, addressing three stringent constraints of resource-limited deployment: minimal hardware requirements, offline functionality, and robust cross-domain generalization.

UltraSeg-130K (0.13M parameters) achieves real-time inference on commodity single-core CPU (51.7 FPS) and Android mobile devices (34.8 FPS), targeting resource-constrained mobile health settings; UltraSeg-500K (0.5M parameters) maintains real-time performance on CPU (27.8 FPS) and Android devices (16.1 FPS), suited for specialized bedside examinations requiring higher precision\footnote{CPU: Intel i5-14600K (single-core); Android: Samsung Galaxy A53.}.

Through systematic validation across ten publicly available datasets spanning six anatomical sites (breast, thyroid, kidney, carotid artery, fetal assessment, and small-animal tumor), including two external validation sets, we demonstrate that this framework effectively adapts to diverse clinical ultrasound segmentation challenges. Remarkably, UltraSeg matches or exceeds the performance of the 31M-parameter standard UNet and TransUNet. Despite its extreme compression, the model delivers superior boundary delineation accuracy (HD95), significantly outperforming not only comparable lightweight architectures but also heavyweight models such as UNet and Att-UNet. Furthermore, it exhibits superior zero-shot cross-dataset generalization capabilities compared to most existing approaches. By enabling GPU-free deployment on edge devices or standard CPU-based workstations, UltraSeg democratizes access to advanced AI-assisted diagnostics in low-resource healthcare environments.

The core contribution of this study lies in demonstrating the feasibility of coupling low-cost AI with inherently low-cost imaging modalities. By compressing model parameters to the sub-million level and eliminating GPU dependency for clinical-grade segmentation, UltraSeg harmonizes the cost structure of AI-assisted diagnosis with the accessible nature of ultrasound imaging. This carries paradigmatic significance for sustainable deployment in LMICs and fiscally constrained community hospitals: healthcare institutions can autonomously operate high-performance segmentation algorithms on local hardware without investing in expensive AI server infrastructure or cloud services, achieving true local ownership of technology.

Ultrasound is foundational to global primary healthcare, yet most primary care facilities lack diagnostic imaging~\cite{nafade2024value}. Lowering AI deployment thresholds is therefore not merely technical optimization, but a step toward health equity, ensuring patients in resource-limited regions access AI-assisted diagnostics comparable to developed areas. This bridges the gap from device accessibility to diagnostic capability accessibility.

\begin{figure*}[h]
    \centering
    \includegraphics[width=\textwidth]{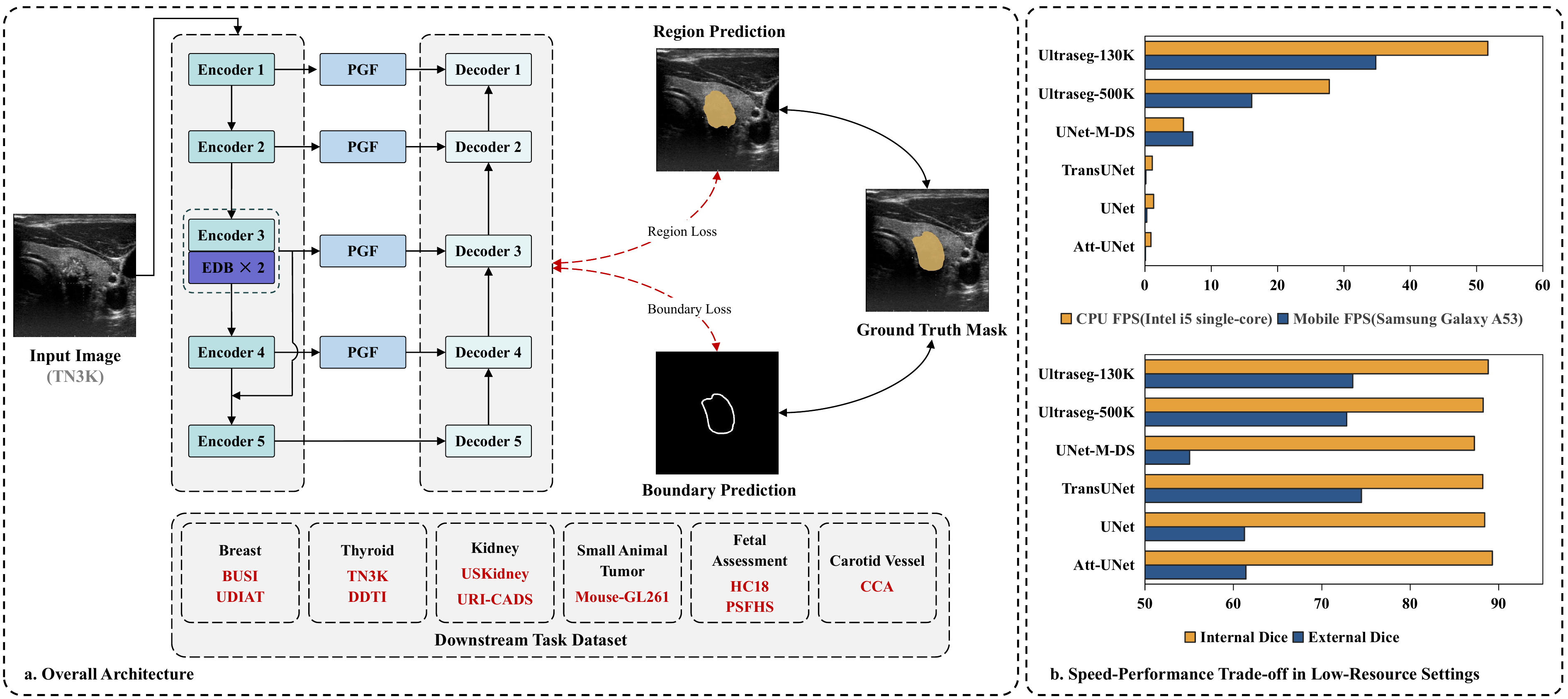} 
    \caption{Overall framework of UltraSeg and comparative benchmarking. \textbf{a}, Architecture overview: UltraSeg employs a compact five-level encoder-decoder topology with reduced channel dimensions. Two cascaded Enhanced Dilated Blocks (EDBs) are integrated at the third encoder stage for multi-scale receptive field expansion, accompanied by attention mechanisms for cross-layer feature refinement. The decoder incorporates dual deep supervision (region and boundary prediction) to strengthen contour delineation. \textbf{b}, Performance comparison of top-performing models, illustrating the trade-off between inference speed (FPS on single-core CPU and mobile devices) and segmentation accuracy across eight internal datasets and two external validation sets.}
    \label{img:fig1}
\end{figure*}

Figure~\ref{img:fig1} illustrates the overall architectural design and comparative performance. Through validation across six segmentation tasks and rigorous external generalization testing, this study confirms that accurate ultrasound segmentation can be efficiently achieved via ultra-lightweight architectures, offering a clinically viable pathway for AI deployment in resource-constrained environments. Source code is publicly available to ensure reproducibility and facilitate clinical adoption.

\section{Method}
\label{sec:method}

\subsection{Overview of UltraSeg architecture}

UltraSeg is a general-purpose lightweight segmentation architecture designed for resource-constrained clinical environments, initially validated on colonoscopic polyp segmentation tasks~\cite{gao2026enabling}. Its core design adheres to the parametric minimalism principle: achieving representation capabilities comparable to large models through architectural prior design rather than post-hoc compression, operating under the constraint parameters.

The architecture adopts an Encoder-Decoder topology. Addressing the characteristics of medical imaging where lesion boundaries are ambiguous and morphologies vary across scales, it constructs three hierarchical computational units:

\textbf{Multi-scale Context Aggregation:} Through the Enhanced Dilated Block (EDB), parallel dilated convolutions are injected at the third encoding layer, expanding the receptive field at a negligible parameter cost. This effectively captures scale variations in medical images.

\textbf{Cross-layer Feature Fusion:} Dual mechanisms of Attention-Guided Fusion (AGF) and Simple Spatial Attention (SSA) are employed. AGF adaptively fuses shallow details and deep semantics via spatial masks. SSA introduce channel-wise competition and recalibration at the bottleneck layer.

\textbf{Global context modeling:} Predictive Gated Fusion (PGF) and Group Shuffle Attention (GSA) are integrated into the upsampling stages of the U-shaped architecture. PGF introduces learnable region-boundary weights that enhance edge perception, while GSA reduces model parameters and improves structural generalization through efficient channel grouping and shuffling operations within the decoder layers. Additionally, the architecture employs dual deep supervision on regions and boundaries, jointly optimizing Dice loss and boundary loss to enhance sensitivity to lesion contours.

In this study, we select two configurations from the family to validate their generalizability in ultrasound scenarios: \textbf{UltraSeg-130K} (0.13M parameters, targeting mobile real-time inference) and \textbf{UltraSeg-500K} (0.5M parameters, targeting high-precision workstation deployment). Both maintain identical topological structures, achieving capacity scaling solely through adjustment of the channel schedule, ensuring performance scalability from extreme lightweight to medium parameter regimes.

\subsection{Adaptive Receptive Field Extension for Ultrasound}
\label{sec:methods_edb}

To accommodate ultrasound lesions that typically exhibit larger field-of-view occupancy and are located in deep tissue, we adjusted the dilation rates of the EDB. The parallel branch dilation rates were modified from the colonoscopy configuration $\left[1,2,3\right]$ to $\left[2,4,6\right]$, expanding the receptive field coverage while maintaining the grouped convolution structure and identical parameter budget.

Specifically, for a $3\times3$ kernel, the receptive field is calculated as $RF = 1 + 2(k-1)d$. The original configuration $\left[1,2,3\right]$ provided receptive fields of $3\times3$, $5\times5$, and $7\times7$ pixels at the feature-map scale, whereas the new configuration $\left[2,4,6\right]$ extends these to $5\times5$, $9\times9$, and $13\times13$ pixels. Given that EDBs are deployed at the third encoding stage (downsampling stride of 4), two cascaded EDB blocks achieve an effective receptive field in the original image space expanding from approximately 50 pixels to approximately 100 pixels, sufficient to cover the complete context of large ultrasound lesions and their surrounding tissue characteristics.

This adaptation is achieved with zero additional parameter overhead, adhering to the lightweight domain adaptation design principle. Its effectiveness is validated in the ablation study.

\subsubsection{Loss function}

To address the inherent \textbf{low-contrast boundaries} and \textbf{speckle noise} in ultrasound imaging, we adopt the region-boundary joint deep supervision framework from the original UltraSeg.

The total loss combines three components:
\begin{equation}
\mathcal{L}_{\text{total}} = \mathcal{L}_{\text{region}} + \mathcal{L}_{\text{boundary}} + \mathcal{L}_{\text{gt}}
\end{equation}

where:
\begin{itemize}
    \item $\mathcal{L}_{\text{gt}}$: Final segmentation loss using BceDiceLoss.
    \item $\mathcal{L}_{\text{boundary}}$: Deep supervision on boundary predictions at three decoder levels with depth-increasing weights $(0.1, 0.2, 0.3)$. This explicit boundary constraint compensates for gradient vanishing at weak ultrasound boundaries where echogenicity transitions are subtle.
    \item $\mathcal{L}_{\text{region}}$: Intermediate region supervision at four levels with weights $(0.1, 0.2, 0.3, 0.4)$, ensuring robust feature aggregation under low signal-to-noise conditions.
\end{itemize}

Unlike optical endoscopy where color contrast aids boundary delineation, ultrasound lesions rely heavily on geometric contour cues. The boundary branch thus provides critical gradient signals for maintaining segmentation topology when intensity gradients are ambiguous. This configuration requires \textit{zero additional parameters} while enhancing robustness to ultrasound-specific artifacts through multi-task learning.

\subsubsection{Architectural adaptation for dual-class ROI segmentation}

For dual-class ROI segmentation scenarios in obstetric ultrasound (e.g., simultaneous segmentation of fetal head and pubic symphysis), we introduce task-specific modifications to the UltraSeg architecture.

\textbf{Multi-Class Adaptive PGF.} The original Predictive Gated Fusion assumes single-channel binary predictions. To accommodate dual-class tasks, we extend it to support dynamic channel processing: Sigmoid activation is applied for single-channel inputs, while for multi-channel inputs (3 channels: background, ROI-A, ROI-B), Softmax normalization is first applied, followed by taking the maximum probability across foreground classes as the region saliency map. This ensures correct broadcasting with skip connection features, enabling seamless compatibility with both single-class and multi-class tasks without additional parameters.

\textbf{Hierarchical Asymmetric Deep Supervision.} Leveraging the physical characteristics of ultrasound imaging where deep layers encode clear semantics while shallow layers suffer from speckle noise, we adopt a \textbf{deep semantics, shallow localization} strategy:
\begin{itemize}
    \item \textbf{Deep layers (Decoder 5, 4):} Output 3-channel logits, utilizing large receptive fields ($\sim$100 pixels) to capture global anatomical context, focusing on semantic discrimination (e.g., fetal head \textit{vs.} pubic symphysis);
    \item \textbf{Shallow layers (Decoder 3, 2):} Output 2-channel logits, performing foreground/background binary classification only, avoiding speckle noise interference on fine-grained discrimination.
\end{itemize}

\textbf{Unified Boundary Supervision.} Despite differing ROI categories, all foreground regions share geometric boundary characteristics. The boundary prediction head maintains single-channel output, with unified edge labels extracted from fused foreground masks via the Canny operator, supervising jointly with region predictions. Deep multi-class predictions provide semantic guidance to shallow binary classification through PGF, ensuring correct categorization of boundary pixels while shallow predictions refine spatial localization. This adaptation incurs a modest parameter increase for multi-class tasks (detailed in Supplementary Document).

\subsection{Training and inference settings}

The training pipeline employed standard data augmentation techniques, including random rotation, scaling, and elastic deformation, implemented within a PyTorch-based framework. All models were trained from scratch on each dataset using the Adam optimizer with a batch size of 4. Input images were uniformly resized to 256×256  pixels. To prevent data leakage, distinct data splitting strategies were adopted for different datasets; detailed configurations are provided in the supplementary materials. Results are reported as the mean across three independent runs initialized with different random seeds. For external validation, predictions were generated by ensemble averaging the outputs from the three trained checkpoints.

\section{Experiments and results}
\label{sec:experiment}

\subsection{Datasets and evaluation metrics}

This study conducts systematic validation on ten publicly available authoritative ultrasound segmentation datasets, encompassing five prevalent human clinical scenarios (breast lesions, thyroid nodules, carotid arteries, kidney structures, and obstetric imaging) along with one biomedical experimental scenario for small-animal tumor assessment (Table~\ref{tab:datasets}). To ensure the clinical reliability of model evaluation, all datasets adhere to the patient-level split principle to the best of our knowledge, minimizing data leakage by strictly enforcing no subject overlap between training and test sets.

Specific splitting strategies are tailored to the characteristics of each dataset. We adopt an image-level random 8:2 split for BUSI~\cite{ALDHABYANI2020104863}, URI-CADS~\cite{molina2024uri}, and USKidney~\cite{song2022ct2us,kucs2024medsegbench}. TN3K follows the official standard splitting protocol~\cite{9434087}; the Common Carotid Artery dataset is partitioned chronologically~\cite{momot2022common}; for obstetric scenarios, HC18~\cite{van2018automated} and PSFHS~\cite{chen2024psfhs} implement an 8:2 split based on fetal ID; and the Mouse-GL261 dataset is partitioned by individual animals~\cite{dorosti2025high}. For each task a validation subset is further derived to enable early stopping during model training. The specific partitioning approach for each dataset is detailed in the Supplementary Document. Detailed statistics, image resolution distributions, and splitting strategies for each dataset are also provided therein.

Specifically, the breast segmentation dataset UDIAT~\cite{yap2017automated} and the thyroid segmentation dataset DDTI~\cite{pedraza2015open} are strictly reserved as independent external validation sets to evaluate the cross-device and cross-center generalization performance of models trained on BUSI and TN3K, respectively.

Regarding data preprocessing, the original HC18 dataset provides only ellipse annotations suitable for fetal skull circumference measurement. We convert these to pixel-wise segmentation masks through ellipse region filling to meet the training requirements of segmentation networks. Furthermore, PSFHS constitutes an obstetric dual-class simultaneous segmentation task requiring concurrent delineation of two distinct anatomical structures, the Pubic Symphysis and Fetal Head, with heterogeneous shapes and potential spatial adjacency. This presents higher complexity than binary foreground/background classification, as the model must disambiguate between two foreground classes without explicit region prompts. All remaining datasets are single-class ROI segmentation tasks.

\begin{table*}[h]
\centering
\caption{Overview of datasets used in this study.}
\label{tab:datasets}
\resizebox{0.9\textwidth}{!}{%
\begin{tabular}{@{}lcccc@{}}
\toprule
\textbf{Dataset} & \textbf{Task} & \textbf{Images} & \textbf{Target} & \textbf{Usage} \\ \midrule
BUSI~\cite{ALDHABYANI2020104863} & Breast Lesion Seg. & 780 & Binary & Development \\
TN3K~\cite{9434087} & Thyroid Nodule Seg. & 3,493 & Binary & Development \\
CCA~\cite{momot2022common} & Carotid Vessel Segmentation & 1,100 & Binary & Development \\
HC18~\cite{van2018automated} & Fetal Head Seg.$^*$ & 999 & Binary & Development \\
USKidney~\cite{song2022ct2us,kucs2024medsegbench} & Kidney Segmentation & 4,586 & Binary & Development \\

URI-CADS~\cite{molina2024uri} & Kidney Segmentation & 1,985 & Binary & Development \\

PSFHS~\cite{chen2024psfhs} & Pubic Symphysis \& Fetal Head & 4,074 & 2-class simultaneous & Development \\
Mouse-GL261~\cite{dorosti2025high} & C57BL/6 Mice Brain Glioma Seg. & 1,856 & Binary & Development \\ 
\midrule
UDIAT~\cite{yap2017automated} & Breast Lesion Seg. & 163 & Binary & External Test \\
DDTI~\cite{pedraza2015open} & Thyroid Nodule Seg. & 637 & Binary & External Test \\ 
\bottomrule
\multicolumn{5}{l}{\footnotesize $^*$Ellipse annotations converted to pixel-wise masks.}
\end{tabular}%
}
\end{table*}

In this study, Dice served as the primary evaluation metric, with IoU and HD95 as supplementary measures. Dice and IoU quantify volumetric agreement (range 0–1, higher is better), while HD95 measures boundary delineation accuracy (lower values indicate superior edge localization).

For the Mouse-GL261 dataset, we do not report HD95 due to the small sample size (n=12) and the metric's sensitivity to outliers. Frequent segmentation failures in this dataset can produce unbounded HD95 values, making the statistic unreliable.

\subsection{Benchmarking strategy}

To establish a rigorous evidence base for clinical deployment scenarios spanning diverse resource constraints, we conducted a systematic performance evaluation across a five-order-of-magnitude spectrum of model complexities (from 38K to 105M parameters). Our benchmarking framework spans distinct deployment categories:

\textbf{(1) High-resource clinical workstations:} Standard U-Net (31M)~\cite{ronneberger2015u}, Attention U-Net (34M)~\cite{oktay2018attention}, and TransUNet (105M)~\cite{chen2024transunet} represent current server-grade deployment standards. These establish the performance ceiling against which lightweight alternatives are judged.

\textbf{(2) Moderate-compression architectures:} U-Net++ (4.9M)~\cite{zhou2019unetpp}, UNeXt (1.6M)~\cite{valanarasu2022unext}, and scaled U-Net variants using DSConv (0.39M--1.5M) explore the middle ground between accuracy and computational efficiency, typical of modern GPU-equipped ultrasound devices.

\textbf{(3) Edge-deployment candidates:} FastSCNN (0.37M)~\cite{poudel2019fast}, MobileUNet (0.14M)~\cite{2021Real}, and dermatology-specialized compact models (EGE-UNet: 0.053M; LB-UNet: 0.038M)~\cite{ruan2023ege,xu2024lb} represent the current state-of-the-art for mobile and embedded clinical environments.

\textbf{(4) Proposed UltraSeg family:} UltraSeg-500K and UltraSeg-130K complete this systematic validation continuum. These models employ identical architectural topologies with parameterized capacity scaling, enabling direct assessment of the minimal parameter threshold required for clinically reliable ultrasound segmentation across multiple anatomical targets.

All experiments in this study were conducted using the PyTorch framework on two NVIDIA 5090 GPUs. To ensure robustness, each experiment was repeated three times with three fixed random seeds. The experimental settings were uniformly configured as follows: input image resolution of $256\times256$, batch size of 4, and 2 data loading workers. The Adam optimizer was employed with a hierarchical learning rate strategy based on model capacity: a larger learning rate of $1\times10^{-3}$ was used for models with fewer than 5M parameters, while a learning rate of $3\times10^{-4}$ was adopted for models exceeding 5M parameters. Given the substantial parameter range across our experiments (from 0.038M to 105.32M), this tiered learning rate scheduling was implemented to ensure fair and comparable performance evaluation across architectures of vastly different scales. Ablation studies comparing results under different learning rate configurations are provided in the Supplementary Materials. All models were trained for a maximum of 100 epochs, with early stopping triggered if the validation Dice coefficient showed no improvement for 10 consecutive epochs. The final results are reported on the unseen test set. Additionally, although ultrasound images are inherently grayscale, we replicated the single-channel input to three channels to ensure broader architectural compatibility.

\subsection{Main results}

As summarized in Table~\ref{tab:mainresults1}, the segmentation performance of different methods across seven ultrasound datasets is presented. 

For the BUSI dataset, UltraSeg-500K achieved a mean Dice score of 72.33 with only 0.5M parameters, outperforming all competing methods. The smaller-scale UltraSeg-130K also attained a competitive Dice score of 71.49. Across all seven ultrasound segmentation tasks, the UltraSeg family not only significantly outperformed other lightweight architectures but also approached or even exceeded the performance of heavyweight baselines including UNet, Attention U-Net, and TransUNet.

Other lightweight models (excluding UltraSeg) exhibited significant performance degradation in both segmentation accuracy (Dice) and boundary localization (HD95) due to parameter constraints. General-purpose architectures such as UNet++, UNeXt, and FastSCNN demonstrated particularly pronounced boundary errors. Furthermore, even domain-specific lightweight models for medical imaging (e.g., LB-UNet and EGE-UNet), despite employing similar deep supervision mechanisms, consistently underperformed compared to the UltraSeg series across all seven ultrasound segmentation tasks.

\begin{table*}[t]
\centering
\small
\resizebox{\textwidth}{!}{%
\begin{threeparttable}
\caption{Performance comparison on multiple ultrasound datasets. All metrics represent the mean across three independent runs with different random seeds; corresponding standard deviations are provided in the Supplementary Document.}
\label{tab:mainresults1}
\setlength{\tabcolsep}{3pt}
\begin{tabular}{@{}lcccccccccccccccc@{}}
\toprule
\multirow{2}{*}{Method} & \multicolumn{2}{c}{BUSI} & \multicolumn{2}{c}{CCA} & \multicolumn{2}{c}{HC18} & \multicolumn{2}{c}{TN3K} & \multicolumn{2}{c}{USKidney} & \multicolumn{2}{c}{URI-CADS} & \multicolumn{1}{c}{Mouse-GL261$^\dagger$} & \multicolumn{2}{c}{Efficiency} \\
\cmidrule(lr){2-3} \cmidrule(lr){4-5} \cmidrule(lr){6-7} \cmidrule(lr){8-9} \cmidrule(lr){10-11} \cmidrule(lr){12-13} \cmidrule(lr){14-14} \cmidrule(lr){15-16}
& Dice$\uparrow$ & HD95$\downarrow$ & Dice$\uparrow$ & HD95$\downarrow$ & Dice$\uparrow$ & HD95$\downarrow$ & Dice$\uparrow$ & HD95$\downarrow$ & Dice$\uparrow$ & HD95$\downarrow$ & Dice$\uparrow$ & HD95$\downarrow$ & Dice$\uparrow$ & Params$\downarrow$ & FLOPs$\downarrow$ \\
\midrule

TransUNet &  \underline{71.59} & \textbf{33.46} &\textbf{95.26} & \textbf{3.58} & \textbf{97.81} & \textbf{5.41} & \underline{84.73} & \textbf{23.65} & 97.78 & \textbf{6.24} & \textbf{92.82} & \textbf{9.19} & 70.60 & 105.32M & 32.24G \\
Att-UNet & 66.34 & 57.17 & 94.72 & 6.91 & \underline{97.77} & 6.52 & \textbf{84.96} & 28.76 & \underline{97.88} & 6.44 & \underline{92.30} & \underline{10.64} & 75.36 & 34.88M & 66.92G \\
UNet & 70.58 & 53.35 & 94.46 & 7.24 & 97.71 & 7.75 & 84.00 & 30.44 & \textbf{97.91} & 6.76 & 92.21 & 10.95 & \underline{75.41} & 31.04M & 54.80G \\
UNet++ & 62.14 & 68.58 & 93.66 & 11.99 & 96.95 & 14.21 & 78.61 & 53.70 & 96.86 & 12.72 & 89.64 & 16.31 & 74.05 & 4.87M & 6.21G \\
UNeXt & 49.88 & 99.45 & 88.30 & 41.80 & 91.40 & 44.24 & 62.03 & 96.81 & 89.23 & 56.76 & 84.12 & 29.13 & 72.06 & 1.61M & 1.66G \\
UNet-M-DS & 67.27 & 53.91 & 93.66 & 6.31 & 97.48 & 8.71 & 81.20 & 36.19 & 97.68 & 7.39 & 90.89 & 12.81 & 75.14 & 1.51M & 3.72G \\
UNet-S-DS & 66.34 & 57.17 & 92.47 & 9.13 & 97.33 & 9.94 & 79.64 & 36.62 & 97.46 & 8.22 & 90.40 & 13.64 & 74.96 & 0.39M & 1.01G \\
FastSCNN & 61.18 & 76.91 & 89.11 & 37.66 & 95.95 & 20.67 & 69.82 & 65.66 & 96.55 & 18.88 & 86.48 & 23.03 & 69.27 & 0.37M & 2.21G \\
MobileUNet & 35.17 & 115.26 & 55.72 & 109.87 & 83.29 & 57.44 & 43.28 & 119.19 & 75.84 & 78.05 & 73.84 & 38.67 & 65.42 & 0.14M & 0.17G \\
EGE-UNet & 62.09 & 71.37 & 88.18 & 9.23 & 96.23 & 13.55 & 75.14 & 46.37 & 96.78 & 10.82 & 85.84 & 20.09 & 71.93 & 0.05M & 0.07G \\
LB-UNet & 65.27 & 64.43 & 93.01 & 7.47 & 96.69 & 9.06 & 79.54 & 46.71 & 97.26 & 7.44 & 89.56 & 16.26 & 72.73 & 0.04M & 0.10G \\
\midrule
UltraSeg-130K & 71.49 & \underline{38.01} & \underline{94.86} &  \underline{3.74} & 97.65 & 5.91 & 83.87 & 29.38 & 97.57 & 6.61 & 90.80 & 12.26 & 75.36 & 0.13M & 0.15G \\
UltraSeg-500K &  \textbf{72.33} & 41.89 & 94.49 & 4.83 & 97.74 & \underline{5.73} & 84.28 & \underline{28.54} & 97.71 & \underline{6.27} & 91.66 & 11.24 & \textbf{77.62} & 0.50M & 0.54G \\
\bottomrule
\end{tabular}
\begin{tablenotes}[flushleft,online]
\footnotesize
\item[$\dagger$] HD95 is not reported due to statistical instability, where high variance in small-sample annotations produced undefined values in at least one run.
\item Bold indicates the best segmentation performance, and underline indicates the second-best.
\end{tablenotes}
\end{threeparttable}%
}
\end{table*}

Table~\ref{tab:mainresults2} presents the zero-shot cross-dataset generalization performance of different models: trained on BUSI and tested on UDIAT (breast domain), and trained on TN3K and tested on DDTI (thyroid domain). The UltraSeg family not only substantially outperformed other lightweight architectures, but also significantly surpassed Att-UNet and UNet.

UltraSeg marginally underperformed TransUNet and Att-UNet on TN3K, yet demonstrated superior generalization when transferred to DDTI. Specifically, Att-UNet exhibited a pronounced failure in HD95, accompanied by substantial Dice degradation, whereas the UltraSeg family maintained consistent performance across both datasets. These results indicate that substantial parameter reduction does not compromise generalization; rather, UltraSeg achieves superior domain transferability by mitigating overfitting, thereby outperforming heavyweight baselines in cross-domain scenarios.

\begin{table*}[t]
\centering
\small
\begin{threeparttable}
\caption{Performance comparison on zero-shot cross-dataset generalization UDIAT and DDTI datasets. Dice is reported in percentage ($\%$).}
\label{tab:mainresults2}
\setlength{\tabcolsep}{8pt}
\begin{tabular}{@{}lcccc@{}}
\toprule
\multirow{2}{*}{Method} & \multicolumn{2}{c}{UDIAT (Trained on BUSI)} & \multicolumn{2}{c}{DDTI (Trained on TN3K)} \\
\cmidrule(lr){2-3} \cmidrule(lr){4-5}
& Dice$\uparrow$ & HD95$\downarrow$ & Dice$\uparrow$ & HD95$\downarrow$ \\
\midrule
TransUNet & \textbf{75.77} & \textbf{25.34} & 73.18 & \textbf{38.06} \\
Att-UNet & 64.65 &36.67 & 58.19 & 56.21 \\
UNet & 64.53 & 44.68 & 57.96 & 50.53 \\
UNet++ & 52.42 & 57.59 & 47.29 & 69.63 \\
UNeXt & 53.22 & 50.22 & 56.87 & 53.02 \\
UNet-Medium-DS & 48.59 & 65.16 & 60.67 & 50.69 \\
UNet-Small-DS & 42.20 & 91.32 & 27.38 & 93.49 \\
FastSCNN & 27.74 & 146.06 & 21.25 & 117.83 \\
MobileUNet & 45.30 & 90.26 & 43.33 & 69.01 \\
EGE-UNet & 68.32 & 57.76 & 64.11 & 66.64 \\
LB-UNet & 51.21 & 81.11 & 60.33 & 52.33 \\
\midrule
UltraSeg-130K & \underline{73.34} & \underline{35.43} & \textbf{73.66} & \underline{41.21} \\
UltraSeg-500K & 72.38 & 35.92 & \underline{73.23} & 41.50 \\
\bottomrule
\end{tabular}
\begin{tablenotes}[]
\footnotesize \item Bold indicates the best segmentation performance, and underline indicates the second-best.
\end{tablenotes}
\end{threeparttable}
\end{table*}

In obstetric imaging (fetal ultrasound examination), we employed two authoritative public datasets. The HC18 dataset focuses on single-class ROI segmentation, specifically fetal head segmentation to support biometric measurement~\cite{van2018automated}, while the PSFHS dataset addresses dual-class ROI simultaneous segmentation, requiring concurrent delineation of the fetal head and pubic symphysis~\cite{chen2024psfhs}. The original UltraSeg family was designed exclusively for single-class ROI segmentation; we therefore performed architectural fine-tuning to extend its capability to multi-class scenarios.

Table~\ref{tab:mainresults3} presents the performance comparison of various models on the PSFHS dataset. LB-UNet and EGE-UNet are excluded from this comparison as their architectures are restricted to single-class ROI segmentation. The results demonstrate that the UltraSeg family achieved mean Dice scores of 94.39 (130K variant) and 94.74 (500K variant), significantly outperforming other lightweight models while approaching the performance of large-scale architectures with over 30M parameters. Regarding the HD95 metric, the UltraSeg family surpassed all comparison models except TransUNet, with only marginal inferiority to the latter. This indicates that UltraSeg, with merely 0.13M and 0.5M parameters, delivers more stable boundary prediction performance in the dual-class segmentation task of fetal head and pubic symphysis.

\begin{table*}[t]
\centering
\small
\caption{Performance comparison on PSFHS (Pubic Symphysis \& Fetal Head Segmentation). All metrics represent the mean across three independent runs with different random seeds; corresponding standard deviations are provided in the Supplementary Document.}
\label{tab:mainresults3}
\setlength{\tabcolsep}{8pt}
\begin{tabular}{@{}lccccc@{}}
\toprule
\multirow{2}{*}{Method} & Pubic Symphysis & Fetal Head & \multicolumn{2}{c}{Mean} \\
\cmidrule(lr){2-2} \cmidrule(lr){3-3} \cmidrule(lr){4-5}
& Dice$\uparrow$ & Dice$\uparrow$ & Dice$\uparrow$ & HD95$\downarrow$ \\
\midrule
TransUNet & \underline{93.62} & \textbf{96.49} & \textbf{95.06} & \textbf{8.11} \\
Att-UNet & 93.60 & \underline{96.40} & 95.00 & 8.46 \\
UNet & \textbf{93.67} & \underline{96.40} & \underline{95.03} & 8.74 \\
UNet++ & 92.51 & 95.46 & 93.98 & 14.84 \\
UNeXt & 90.75 & 91.19 & 90.97 & 34.45 \\
UNet-Medium-DS & 93.24 & 96.13 & 94.68 & 9.92 \\
UNet-Small-DS & 92.75 & 96.00 & 94.37 & 10.35 \\
FastSCNN & 91.58 & 95.00 & 93.29 & 14.94 \\
MobileUNet & 75.43 & 79.63 & 78.33 & 55.33 \\
\midrule
UltraSeg-130K & 92.57 & 96.22 & 94.39 & 8.38 \\
UltraSeg-500K & 93.18 & 96.31 & 94.74 & \underline{8.26} \\
\bottomrule
\multicolumn{5}{@{}l@{}}{\footnotesize Bold indicates the best segmentation performance, and underline indicates the second-best. }
\end{tabular}
\end{table*}

For the breast lesion segmentation tasks involving the BUSI dataset and the external validation set UDIAT, both datasets provide not only pixel-level segmentation masks but also pathology labels indicating benign or malignant status. Table~\ref{tab:subgroup} presents the stratified segmentation performance of the UltraSeg family against representative large models (UNet and TransUNet) across these subgroups.

On BUSI, UltraSeg models exhibited slightly reduced performance for benign samples compared to heavyweight architectures, yet demonstrated substantial superiority in malignant lesion segmentation. Specifically, UltraSeg-500K achieved a Dice score 2.2 percentage points higher than TransUNet on malignant samples. 

Regarding external validation on UDIAT, UltraSeg remained marginally below TransUNet on benign cases but outperformed baseline UNet by 7.29 Dice points. For malignant samples, UltraSeg consistently surpassed UNet, though remaining slightly below TransUNet.

These results on BUSI and the zero-shot inference on UDIAT demonstrate that the proposed CNN-based UltraSeg architecture, trained from scratch without reliance on ImageNet pre-training, can match or exceed the performance of GPU-dependent large models exceeding 31M parameters, while maintaining particular robustness in segmenting malignant lesions.

\begin{table*}[t]
\centering
\small
\resizebox{\textwidth}{!}{%
\begin{threeparttable}
\caption{Benign \textit{vs.} malignant subgroup performance on breast ultrasound. Metrics denote mean Dice ($\%$) across three checkpoints (distinct random seeds) trained on BUSI and evaluated zero-shot on UDIAT.}
\label{tab:subgroup}
\setlength{\tabcolsep}{5pt}
\begin{tabular}{@{}l c cc cc cc cc@{}}
\toprule
& & \multicolumn{4}{c}{\textbf{BUSI\tnote{$\dagger$}}} & \multicolumn{4}{c}{\textbf{UDIAT External\tnote{$\ddagger$}}} \\
\cmidrule(lr){3-6} \cmidrule(lr){7-10}
& & \multicolumn{2}{c}{Benign (n=92)} & \multicolumn{2}{c}{Malignant (n=40)} & \multicolumn{2}{c}{Benign (n=109)} & \multicolumn{2}{c}{Malignant (n=54)} \\
\cmidrule(lr){3-4} \cmidrule(lr){5-6} \cmidrule(lr){7-8} \cmidrule(lr){9-10}
\textbf{Method} & \textbf{Params} & Dice & HD95 & Dice & HD95 & Dice & HD95 & Dice & HD95 \\
\midrule
TransUNet & 105.32M & 77.75 & 24.17 & 70.66 & 54.58 & 80.57 & 21.21 & 75.55 & 32.61 \\
UNet & 31.04M & 76.77 & 48.65 & 67.51 & 63.01 & 71.70 & 43.06 & 67.90 & 47.62 \\
UltraSeg-130K & 0.13M & 74.60 & 30.89 & 72.08 & 54.79 & 78.56 & 32.48 & 72.46 & 39.89 \\
UltraSeg-500K & 0.50M & 74.90 & 36.22 & 72.87 & 55.62 & 78.99 & 34.19 & 71.47 & 41.53 \\
\bottomrule
\end{tabular}
\begin{tablenotes}[flushleft]
\footnotesize
\item[$\dagger$] BUSI subgroup metrics aggregate validation and test sets to ensure sufficient sample size for stratified analysis.
\item[$\ddagger$] The optimal checkpoints trained using BUSI, and inference under zero samples.
\end{tablenotes}
\end{threeparttable}
}
\end{table*}

\begin{figure}[!htbp]  
\centering
\begin{minipage}{0.48\linewidth}
  \centering
  \includegraphics[width=\linewidth]{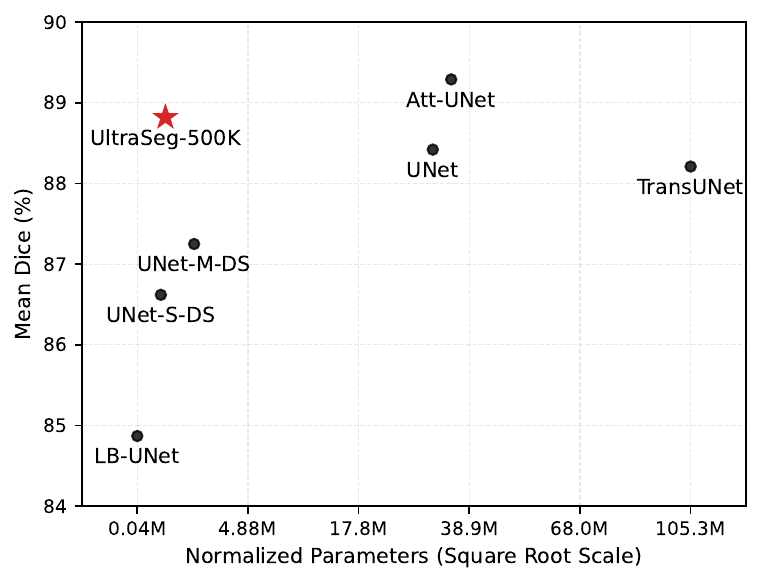}
  \small (a) Parameters vs. Dice
\end{minipage}\hfill
\begin{minipage}{0.48\linewidth}
  \centering
  \includegraphics[width=\linewidth]{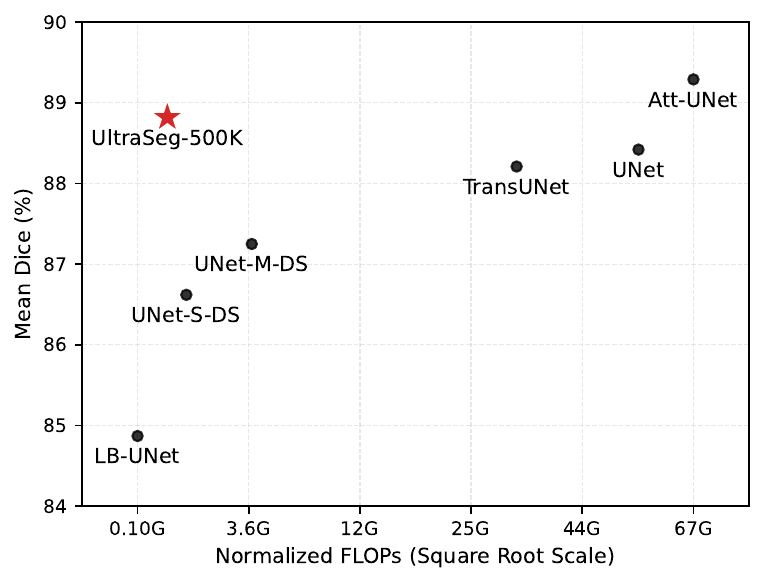}
  \small (b) FLOPs vs. Dice
\end{minipage}

\vspace{10pt}  

\begin{minipage}{0.9\linewidth}  
  \centering
  \includegraphics[width=\linewidth]{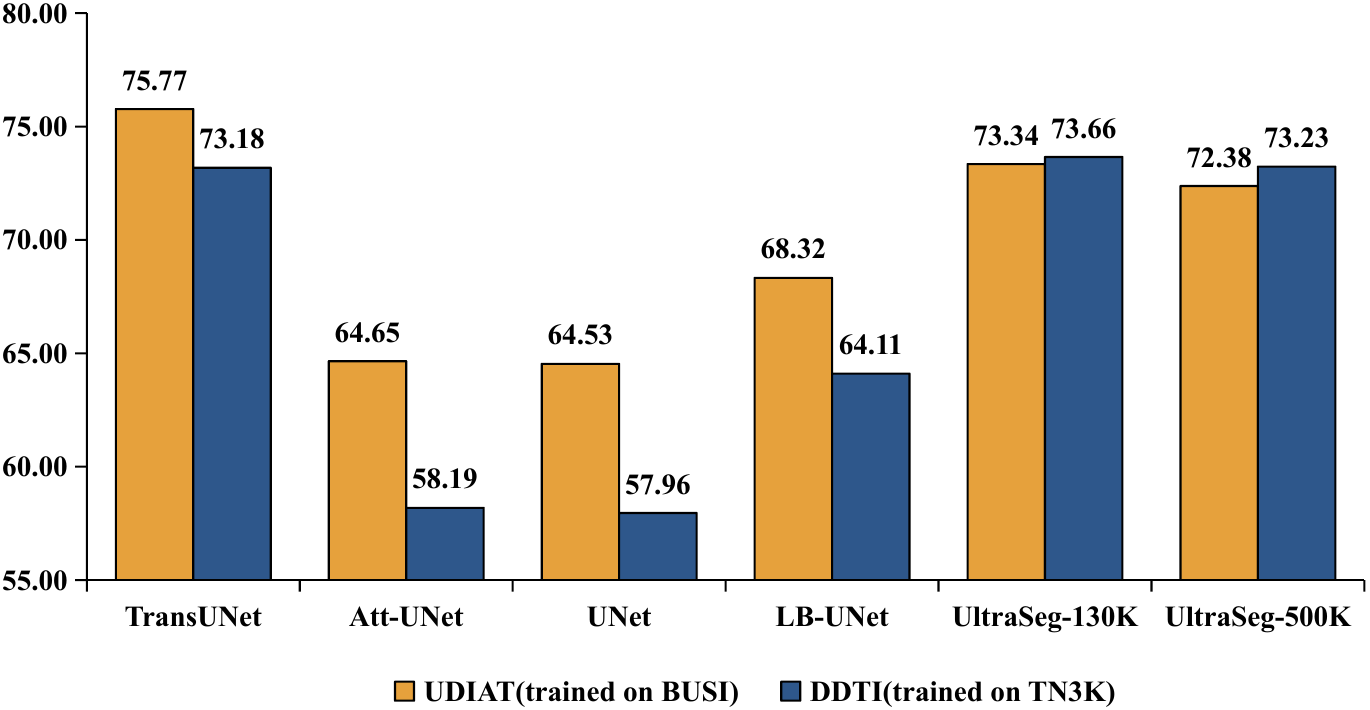}
  \small (c) Cross-dataset generalization on external validation sets
\end{minipage}

\caption{
\textbf{Efficiency-accuracy trade-off and generalization capability.} 
(a) Parameter-Dice and (b) FLOPs-Dice scatter plots demonstrate that UltraSeg-500K (red star) simultaneously achieves the best accuracy and superior efficiency compared to existing methods. 
(c) Zero-shot external validation performance on UDIAT (trained on BUSI) and DDTI (trained on TN3K).
}
\label{fig:tradegen}
\end{figure}

Figure~\ref{fig:tradegen}a illustrates the relationship between parameter count and mean Dice performance across eight internal datasets, while Figure~\ref{fig:tradegen}b compares FLOPs against Dice scores. Given the substantial range of model parameters, both plots employ square-root scaling on the horizontal axes. The UltraSeg-500K achieves an optimal balance between model compactness and segmentation accuracy within the lightweight regime.

Figure~\ref{fig:tradegen}c demonstrates zero-shot generalization performance on two external validation datasets. UltraSeg marginally underperforms TransUNet on BUSI but exhibits slight superiority on DDTI, while substantially outperforming all other comparative models.

\subsection{Inference speed on CPU or bedside device}

To validate the deployment feasibility of the UltraSeg family in resource-limited settings, we conducted systematic evaluations of inference efficiency on edge computing devices. Specifically, we utilized a refurbished Samsung Galaxy A53 (acquired at approximately USD \$130 from the secondary market) as the test platform, representative of hardware accessible to primary care clinics in low-resource environments; additionally, we tested on an Intel i9-14900K single-core CPU to assess performance on standard office equipment or dedicated edge computing nodes.

As illustrated in Table~\ref{tab:efficiency}, UltraSeg achieves a superior trade-off between computational efficiency and segmentation accuracy compared to existing approaches. In the single-core CPU scenario, UltraSeg-130K and UltraSeg-500K achieved inference speeds of 51.7 FPS and 27.9 FPS, respectively; on the same mobile device, both models maintained 34.8 FPS and 16.1 FPS, enabling responsive real-time ultrasound guidance. By contrast, while TransUNet attains competitive segmentation performance on both internal and external validation sets, its prohibitive computational overhead renders it infeasible for cost-effective real-time ultrasound inference; other lightweight competitors (EGE-UNet, LB-UNet), despite their compact model sizes, exhibit significant deficiencies in segmentation accuracy (Internal/External Dice) and fail to achieve real-time performance on single-core CPUs. These results establish UltraSeg as the only solution to our knowledge that simultaneously delivers clinical-grade accuracy and cross-platform real-time inference within the sub-million parameter regime, providing a technically feasible pathway for intelligent ultrasound-assisted diagnosis in primary care institutions lacking GPU infrastructure.

\begin{table*}[t]
\centering
\small
\caption{Comparison of model complexity, segmentation accuracy, and inference throughput.}
\label{tab:efficiency}
\begin{threeparttable}
\resizebox{\textwidth}{!}{%
\begin{tabular}{@{}lccccccc@{}}
\toprule
& & & \multicolumn{2}{c}{\textbf{Accuracy (Dice$\uparrow$)}} & \multicolumn{2}{c}{\textbf{Throughput (FPS$\uparrow$)}} \\
\cmidrule{4-5} \cmidrule{6-7}
Method & Params$\downarrow$ & FLOPs$\downarrow$ & Internal & External & CPU & Bedside \\
\midrule
TransUNet & 105.32M & 32.24G & 88.21 & 74.48  & 1.1 & 0.12 \\
Att-UNet & 34.88M & 66.92G & 89.29   & 61.42  & 0.8 & 0.06 \\
UNet & 31.04M & 54.80G & 88.42   & 61.25  & 1.3 & 0.27 \\
UNet++ & 4.87M & 6.21G & 85.74    & 49.86  & 11.6 & 3.3 \\
UNeXt & 1.61M & 1.66G & 78.50   & 34.79 & 28.7 & 14.2 \\
UNet-M-DS & 1.51M & 3.72G & 87.25    & 55.05  & 5.8 & 7.2 \\
UNet-S-DS & 0.39M & 1.01G & 86.62     & 54.63 & 17.6 & 20.0 \\
FastSCNN & 0.37M & 2.21G & 82.70  & 44.32 & 38.0 & 13.2 \\
MobileUNet & 0.14M & 0.17G & 63.86   & 24.49  & 275.3 & 143.2 \\
EGE-UNet\tnote{$\dagger$} & 0.05M & 0.07G & 82.31  & 55.77  & 84.1 & 38.3 \\
LB-UNet\tnote{$\dagger$} & 0.04M & 0.10G & 84.87  & 66.22 & 57.7 & 41.1 \\
\midrule
\textbf{UltraSeg-130K} & \textbf{0.13M} & \textbf{0.15G} & \textbf{88.25} & \textbf{73.50} & \textbf{51.7} & \textbf{34.8} \\
\textbf{UltraSeg-500K} & \textbf{0.50M} & \textbf{0.54G} & \textbf{88.82} & \textbf{72.81} & \textbf{27.8} & \textbf{16.1} \\
\bottomrule
\end{tabular}}
\begin{tablenotes}[flushleft]
\footnotesize
\item CPU FPS measured on Intel i5-14600K, single-core. Bedside throughput evaluated on Samsung Galaxy A53, demonstrating cost-effective, scalable AI deployment without specialized digital infrastructure in low-resourced clinical settings.
\item[$\dagger$] EGE-UNet and LB-UNet: single-class design, PSFHS excluded from internal Dice/HD95 calculation. External validation datasets: UDIAT (breast), DDTI (thyroid).
\end{tablenotes}
\end{threeparttable}
\end{table*}

\subsection{Qualitative analysis}
\label{bnexplain}

\begin{figure*}[h]
    \centering
    \includegraphics[width=\textwidth]{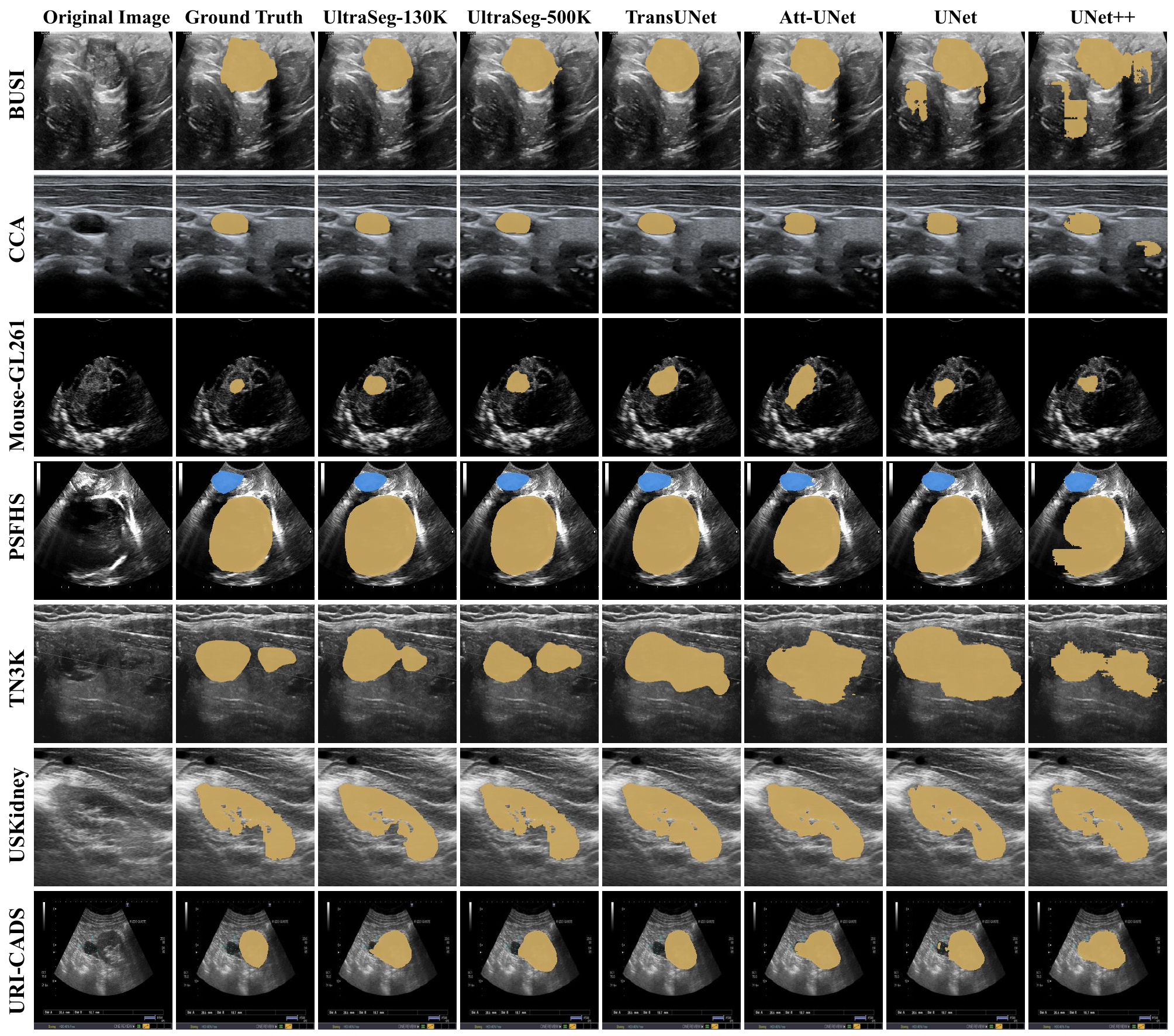} 
    \caption{Qualitative comparisons on different models.}
    \label{img:visual1}
\end{figure*}

Figure~\ref{img:visual1} presents qualitative segmentation results across seven diverse ultrasound datasets. On datasets with relatively homogeneous anatomical structures, such as the Common Carotid Artery (CCA) dataset, UltraSeg exhibits superior boundary delineation compared to Att-UNet, UNet, and UNet++. Similarly, on the USKidney dataset, characterized by complex internal architectures including significant cavities and echogenic interfaces, UltraSeg models demonstrate markedly enhanced capacity to resolve internal structural heterogeneities compared to TransUNet and alternative architectures. On the real-world kidney segmentation dataset URI-CADS, our method demonstrates slightly inferior performance compared to TransUNet, yet exhibits markedly superior edge prediction capability relative to other models.

For clinically challenging datasets such as BUSI and TN3K, UltraSeg maintains robust performance. For example, in a representative case from the TN3K dataset containing two thyroid nodules of disparate sizes, UltraSeg successfully segments these as distinct entities, whereas heavyweight architectures including TransUNet erroneously merge them into a single region.

Furthermore, UltraSeg demonstrates consistent stability on the small-sample Mouse-GL261 dataset and the multi-class PSFHS dataset, underscoring its reliability across varying data scales and task complexities.

\begin{figure*}[h]
    \centering
    \includegraphics[width=\textwidth]{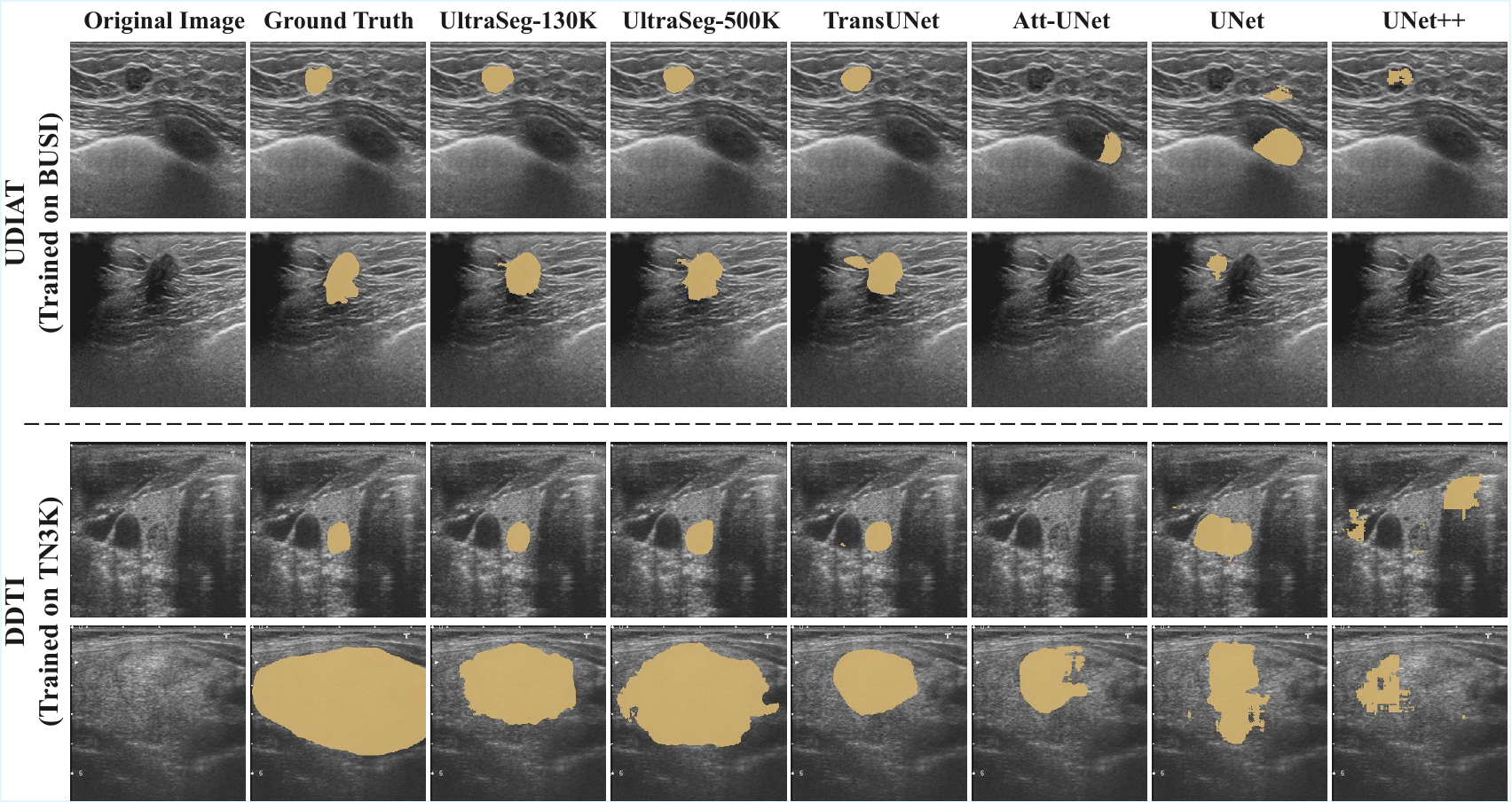} 
    \caption{Comparison of zero-shot cross-dataset inference performance.}
    \label{img:visual2}
\end{figure*}

Figure~\ref{img:visual2} illustrates zero-shot cross-dataset performance on UDIAT and DDTI. Thyroid and breast segmentation are particularly challenging in ultrasound analysis; zero-shot evaluation thus serves as a rigorous test of generalization. As shown, our method significantly outperforms Att-UNet, UNet, and U-Net++. On DDTI, UltraSeg exhibits superior spatial judgment compared to TransUNet, avoiding over-segmentation of small structures while achieving more accurate delineation of large targets.

\subsection{Ablation study}
\label{ablation}

Motivated by the observation that ultrasound anatomical structures are typically larger and deeper than colonoscopy polyps, we expanded the receptive field in the third downsampling layer. To validate the architectural contribution of the Enhanced Dilated Block (EDB) and its task-specific adaptation for ultrasound imaging, we conducted systematic ablation experiments on UltraSeg-130K and UltraSeg-500K (Table~\ref{tab:edb_ablation}). Three configurations were compared: (1)~\textbf{w/o EDB}: the baseline architecture with the EDB module entirely removed; (2)~\textbf{EDB-original}: the default configuration with dilation rates [1,2,3] optimized for colonoscopy (small, superficial lesions); and (3)~\textbf{EDB-ultrasound}: the modified configuration with dilation rates [2,4,6] specifically designed to capture larger receptive fields for deep-seated anatomical structures in ultrasound.

Results demonstrate that removing EDB causes significant performance degradation (Dice drops by $1.3$--$3.1$\%), confirming its essential role in multi-scale feature aggregation. Moreover, the ultrasound-specific dilation settings consistently outperform the original colonoscopy configuration, particularly on BUSI (Dice gains of $1.7$\% for 130K and $0.5$\% for 500K), validating the necessity of task-specific receptive field design across different imaging domains.

\begin{table*}[t]
\centering
\small
\caption{Ablation study on the Enhanced Dilated Block (EDB). Best results for each model are \textbf{boldfaced}. All metrics represent the mean across three independent runs with different random seeds.}
\label{tab:edb_ablation}
\begin{tabular}{@{}llcccc@{}}
\toprule
\multirow{2}{*}{\textbf{Model}} & \multirow{2}{*}{\textbf{EDB Config}} & \multicolumn{2}{c}{\textbf{BUSI}} & \multicolumn{2}{c}{\textbf{TN3K}} \\
\cmidrule(lr){3-4} \cmidrule(lr){5-6}
 & & Dice$\uparrow$ & HD95$\downarrow$ & Dice$\uparrow$ & HD95$\downarrow$ \\
\midrule
\multirow{3}{*}{UltraSeg-130K} 
 & w/o EDB & 68.41 & 53.34 & 82.53 & 35.08 \\
 & EDB-original & 69.76 & 46.55 & 82.93 & 30.52 \\
 & EDB-ultrasound & \textbf{71.49} & \textbf{38.01} & \textbf{83.87} & \textbf{29.38} \\
\midrule
\multirow{3}{*}{UltraSeg-500K} 
 & w/o EDB & 70.90 & 44.96 & 82.64 & 33.51 \\
 & EDB-original & 71.83 & 44.53 &83.49 & 29.76 \\
 & EDB-ultrasound & \textbf{72.33} & \textbf{41.89} &  \textbf{84.28} & \textbf{28.54} \\
\bottomrule
\end{tabular}
\end{table*}

The parameter discrepancy between UltraSeg-130K and UltraSeg-500K arises from their distinct channel configurations. To investigate the impact of model capacity scaling, we expanded the channel dimensions to [48, 96, 288, 384, 576], yielding UltraSeg-4.38M (4.38M parameters), as detailed in Table~\ref{tab:ablation-capacity}. We validated the effects of this channel expansion across four datasets (BUSI, TN3K, USKidney, and PSFHS) using three independent random seeds.

On TN3K, USKidney, and PSFHS datasets, increasing channel capacity within the identical UltraSeg architecture progressively narrowed the performance gap with the best-performing baselines; UltraSeg-4.38M achieved near-parity with best performance on USKidney and PSFHS. Conversely, on the BUSI dataset, where UltraSeg-500K already attained optimal performance, further channel expansion resulted in performance degradation rather than improvement. These findings demonstrate that the optimal channel configuration for the UltraSeg architecture is task-dependent, and larger capacity does not universally guarantee superior performance.

\begin{table*}[t]
\centering
\small
\resizebox{\textwidth}{!}{%

\begin{threeparttable}
\caption{Ablation study on model capacity across different datasets. All metrics represent the mean across three independent runs with different random seeds.}
\label{tab:ablation-capacity}
\setlength{\tabcolsep}{4pt}
\begin{tabular}{@{}l r cc cc cc cc @{}}
\toprule
& & \multicolumn{2}{c}{\textbf{BUSI}} & \multicolumn{2}{c}{\textbf{TN3K}} & \multicolumn{2}{c}{\textbf{USKidney}} & \multicolumn{2}{c}{\textbf{PSFHS(mean)}} \\
\cmidrule(lr){3-4} \cmidrule(lr){5-6} \cmidrule(lr){7-8} \cmidrule(lr){9-10} \cmidrule(lr){7-8}
\textbf{Method} & \textbf{Params$\downarrow$} & Dice$\uparrow$ & $\Delta$ Best (\%) & Dice$\uparrow$ & $\Delta$ Best (\%) & Dice$\uparrow$ & $\Delta$ Best (\%)  & Dice$\uparrow$ & $\Delta$ Best (\%) \\
\midrule
UltraSeg-130K & 0.13M & 71.49 & -1.16\% & 83.87 & -2.26\%  &97.57&-0.34\% & 94.39 & -0.72\% \\
UltraSeg-500K & 0.50M & \textbf{72.33} & \textbf{0.00\%} & 84.28 & -1.78\% &97.71&-0.20\%& 94.74 & -0.33\% \\
UltraSeg-4.38M & 4.38M & 71.32 & -1.40\% & \textbf{85.81} & \textbf{0.00\%} &97.80&-0.11\%& 94.97 & -0.09\% \\
UNet&  31.04M&70.58&-2.42\%&84.00&-2.10\%& \textbf{97.91}&\textbf{0.00\%}&95.03& -0.03\%\\
TransUNet & 105.32M & 71.59 & -1.02\% & 84.73 & -1.25\% &97.78&-0.13\% & \textbf{95.06} & 0.00\% \\
\bottomrule
\end{tabular}
\begin{tablenotes}[flushleft]
\footnotesize
\item $\Delta$ Best (\%) indicates the relative performance gap to the best Dice score on each dataset. Bold indicates the best segmentation performance.

\end{tablenotes}
\end{threeparttable}}
\end{table*}

\section{Discussion}
\label{sec:discussion}

In this study, we systematically validated UltraSeg, an ultra-lightweight convolutional architecture originally developed for real-time colonoscopic polyp segmentation, across six anatomical sites using ten publicly available datasets. These encompass breast lesions (BUSI, UDIAT), thyroid nodules (TN3K, DDTI), renal structures (USKidney, URI-CADS), carotid arteries (Common Carotid Artery), fetal assessment (HC18, PSFHS), and murine brain tumors (Mouse-GL261). We demonstrate that the UltraSeg family, comprising UltraSeg-130K and UltraSeg-500K, achieves Dice scores comparable to or exceeding those of the 31M-parameter U-Net and the 105M-parameter TransUNet across all tasks, while simultaneously reducing boundary delineation error. Crucially, on the UDIAT and DDTI datasets, which were strictly held out for zero-shot cross-dataset generalization testing, UltraSeg exhibited significantly superior robustness in domain transfer scenarios compared to heavyweight baseline models (Table~\ref{tab:mainresults2}).

These results establish a pivotal conclusion: multiple ultrasound segmentation tasks, particularly single ROI Point-of-Care diagnostics, can be effectively achieved with sub-million scale architectures without compromising clinical-grade accuracy.

The heavyweight segmentation models, despite demonstrating remarkable accuracy in controlled settings, face prohibitive deployment barriers that render them clinically ineffective outside well-funded tertiary care centers. Specifically, architectures such as MedSAM or TransUNet necessitate dedicated GPU workstations or cloud-based inference, with hardware acquisition and maintenance costs often exceeding the price of the ultrasound device itself~\cite{chen2024transunet, 2025Medical}. This creates a paradoxical scenario where the cost of intelligence surpasses that of the medical instrument, effectively excluding the very populations poised to benefit most from AI augmentation—primary care institutions, rural clinics, and vast patient populations in low- and middle-income countries where ultrasound serves as the primary diagnostic modality~\cite{gichoya2025leveraging, ciecierski2022artificial}. By decoupling high-performance segmentation from specialized AI infrastructure, this study directly confronts this inequity, providing a validation of clinically actionable solutions that harmonize the cost structure of AI with the economically accessible reality of ultrasound examinations.

The predicament of intelligent healthcare in resource-limited settings extends beyond funding to include gaps in digital infrastructure, trained personnel, and physical space. UltraSeg addresses these constraints through three design principles. First, hardware minimalism: by achieving real-time inference (>27 FPS) on standard CPUs and performance(>16 FPS) on mobile devices, our framework eliminates the need for GPU servers, high-bandwidth internet, or dedicated IT staff, resources that primary care facilities and mobile health campaigns rarely possess. This enables true local ownership, allowing clinics to autonomously deploy and maintain AI capabilities without reliance on cloud APIs or external technical support. Second, offline autonomy: unlike foundation models (e.g., SAM/MedSAM~\cite{2025Medical,kirillov2023segment}) that demand massive computational resources and interactive prompting, UltraSeg provides fully automated segmentation without network connectivity or physician interaction prompts, ensuring functionality in regions with unreliable electricity or network access. Third, energy optimization: the sub-million parameter design significantly reduces power consumption compared to transformer-based architectures, ensuring compatibility with solar-powered or battery-operated portable ultrasound systems common in low-resource deployment scenarios~\cite{prajwal2025study}.

Beyond infrastructure constraints, ultrasound imaging faces unique human capital challenges due to its operator-dependent nature. In resource-limited regions, the scarcity of experienced sonographers presents a more severe bottleneck than device availability~\cite{ginsburg2023survey,russell2022state}. Our results indicate that UltraSeg's precise boundary delineation capabilities and real-time feedback can effectively reduce inter-operator variability. Notably, in the malignant lesion subgroup analysis (Table~\ref{tab:subgroup}), UltraSeg-500K achieved a Dice score 2.2 percentage points higher than TransUNet for malignant breast lesions, suggesting particular value in high-risk screening scenarios where expert supervision is scarce. This technology democratizes diagnostic capability by delivering expert-level segmentation on low-cost hardware, enabling non-specialist clinicians to perform accurate assessments and reducing referrals to distant urban centers.

For low-resource regions, a portable ultrasound device may be procured through government budgets or international aid, yet the GPU workstations required for AI inference remain unattainable. UltraSeg restructures AI deployment and inference costs from the infrastructure level to the consumable level. This reconfiguration of cost architecture transforms AI from a luxury requiring additional fundraising into a fundamental function bundled with the device.

This study has limitations that warrant consideration for future implementation. First, our validation relies on retrospective public datasets. Prospective clinical trials in actual low-resource settings are needed to confirm effectiveness, robustness, and user acceptance. Second, the framework currently handles static B-mode segmentation of one or two ROIs. Extending to multi-lesion scenarios, dynamic video analysis, or 3D volume reconstruction (such as comprehensive fetal cardiac assessment) would require architectural adaptations while preserving computational efficiency. The architecture is thus a specialized lightweight solution for standardized screening tasks, not a universal fix. For variable class numbers, future work could explore conditional computation or dynamic routing with minimal parameter increase.

Deeper constraints stem from the fragmented nature of ultrasound clinical practice. Unlike the DICOM standardization of CT/MRI, ultrasound acquisition parameters (probe frequency, depth, gain, time-gain compensation) depend heavily on operator experience, and post-processing algorithms vary significantly across vendors. While this study demonstrated model robustness to device variation through cross-device validation, it does not resolve fundamental disparities in diagnostic standards. Thyroid nodule malignancy risk stratification, for instance, differs across versions and institutions worldwide; fetal biometric reference curves likewise vary with population genetics. UltraSeg outputs computational segmentation masks rather than direct diagnostic conclusions; converting these into standardized reports still requires additional standardization layers.

Therefore, the lightweight AI deployment paradigm advocated in this study should be clearly delineated in its scope of application: in scenarios supported by existing public datasets, with relatively unified diagnostic standards, and where single-organ primary screening is the main objective, UltraSeg can provide near-expert-level segmentation assistance at extremely low cost, directly empowering primary screening in resource-limited regions. However, for complex interventional procedures relying on real-time operator decision-making, differential diagnosis of difficult cases requiring multi-modal fusion (ultrasound + elastography + contrast), or emerging applications without established public benchmarks, current technology remains insufficient to independently support clinical decision-making.

Sustained progress in these directions requires systematic engineering efforts beyond isolated technical advances. This encompasses multiple critical dimensions: manufacturers opening standardized data interfaces; clinical societies establishing AI-specific quality control protocols; and collaborative data construction between academia and industry.

We have shown that sub-million parameter models can match or surpass hundred-million-parameter architectures in specific clinical tasks. This shifts the deployment threshold from centralized data centers to dispersed primary care settings. UltraSeg is not merely algorithmic lightweighting, but a deliberate alignment of limited computational resources with clinical needs, rejecting complexity for its own sake.

Centralized AI development increasingly dominates the field, yet lightweight and locally deployable solutions offer an alternative to technological monopolization. Clinically effective AI need not rely on expensive cloud APIs or dedicated servers. By reducing AI-assisted diagnosis to the cost of the ultrasound device itself, rural clinics and underfunded hospitals can access these tools on equal terms. Lower deployment costs enable AI-assisted ultrasound to reach across geographic and economic boundaries, serving those who need it most.

\section*{Declaration of competing interest}

The authors declare that they have no known competing financial interests or personal relationships that could have appeared to influence the work reported in this paper.

\section*{Code availability}
The code for this study is publicly available at https://github.com/AI-thpremed/UltraSeg-UltraSound.

\section*{Acknowledgements}

The authors declare that there are no acknowledgements.

\bibliography{ultraseg}
\bibliographystyle{plain}

\appendix

\end{document}